\documentclass[10pt,twocolumn,letterpaper]{article}

\usepackage{cvpr}              
\usepackage[table]{xcolor}
\usepackage{tikz}
\usepackage{multirow}
\usepackage{algorithm}
\usepackage{algpseudocode}
\usepackage{dblfloatfix}
\usepackage{bm}

\definecolor{cvprblue}{rgb}{0.21,0.49,0.74}
\usepackage[pagebackref,breaklinks,colorlinks,citecolor=cvprblue]{hyperref}

\title{Cloud-Device Collaborative Learning for Multimodal Large Language Models}

\author{
    {\normalsize Guanqun Wang$^{1*}$ \quad Jiaming Liu$^{1*}$ \quad Chenxuan Li$^{1*}$ \quad Junpeng Ma$^{1}$ \quad Yuan Zhang$^{1}$ \quad Xinyu Wei$^{1}$ \quad Kevin Zhang$^{1}$}\\ 
    {\normalsize Maurice Chong$^{1}$ \quad Ray Zhang$^{2}$ \quad Yijiang Liu$^{3}$ \quad Shanghang Zhang$^{1\dag}$}\\
    {\normalsize{$^{1}$National Key Laboratory for Multimedia Information Processing, School of Computer Science,}}\\
    {\normalsize Peking University \quad $^{2}$Shanghai AI Lab \quad $^{3}$Nanjing University}
}

\begin{document}
\maketitle
\renewcommand{\thefootnote}{*}
\footnotetext{These authors contributed equally to this work.}
\renewcommand{\thefootnote}{\dag}
\footnotetext{Corresponding authors: shanghang@pku.edu.cn}
\begin{abstract}
The burgeoning field of Multimodal Large Language Models (MLLMs) has exhibited remarkable performance in diverse tasks such as captioning, commonsense reasoning, and visual scene understanding. However, the deployment of these large-scale MLLMs on client devices is hindered by their extensive model parameters, leading to a notable decline in generalization capabilities when these models are compressed for device deployment. Addressing this challenge, we introduce a Cloud-Device Collaborative Continual Adaptation framework, designed to enhance the performance of compressed, device-deployed MLLMs by leveraging the robust capabilities of cloud-based, larger-scale MLLMs.
Our framework is structured into three key components: a device-to-cloud uplink for efficient data transmission, cloud-based knowledge adaptation, and an optimized cloud-to-device downlink for model deployment. In the uplink phase, we employ an Uncertainty-guided Token Sampling (UTS) strategy to effectively filter out-of-distribution tokens, thereby reducing transmission costs and improving training efficiency. On the cloud side, we propose Adapter-based Knowledge Distillation (AKD) method to transfer refined knowledge from large-scale to compressed, pocket-size MLLMs. Furthermore, we propose a Dynamic Weight update Compression (DWC) strategy for the downlink, which adaptively selects and quantizes updated weight parameters, enhancing transmission efficiency and reducing the representational disparity between cloud and device models. Extensive experiments on several multimodal benchmarks demonstrate the superiority of our proposed framework over prior Knowledge Distillation and device-cloud collaboration methods. Notably, we also validate the feasibility of our approach to real-world experiments.
\end{abstract}    
\section{Introduction}
\label{sec:intro}

In recent years, we have witnessed a proliferation of Large Language Models (LLMs) and Multimodal Large Language Models (MLLMs), with models like GPT4 \cite{openai2023gpt4} demonstrating exceptional performance across various tasks, including visual question answering (VQA) and commonsense reasoning.
These MLLMs, such as Flamingo \cite{Alayrac2022Flamingo} and BLIP-2 \cite{li2023blip2}, empower LLMs with the capability to comprehend and reason about visual scenes.
Due to their large amount of parameters, MLLMs are commonly deployed on cloud servers, demonstrating strong generalization capability. However, their large-scale parameters make it challenging to directly deploy MLLMs on the device, which also limits their practicality.

Since the client device is resource-constrained, MLLMs need to be compressed for the deployment on the device.
The compressed MLLMs indeed demonstrate remarkable performance when the test data distribution closely matches the training data distribution.
However, this assumption encounters significant challenges in real-world scenarios, where non-static environments and distribution shifts are prevalent \cite{wang2020tent, wang2022continual}.
The small-size MLLMs are susceptible to severe performance degradation when confronted with dynamic distribution shifts \cite{RiccardoVolpi2020ContinualAO, wang2022continual, liu2023vida}.
There are two principal challenges: (1) The limited computational capacity of edge devices hinders the ability to perform timely model updates, leading to performance decay when encountering distribution shifts. (2) Compressed models, which have a relatively small capacity, struggle to adapt to continuously changing environments, leading to insufficient generalization ability.

\begin{figure*}[h] 
\centering 
\includegraphics[width=0.85\textwidth]{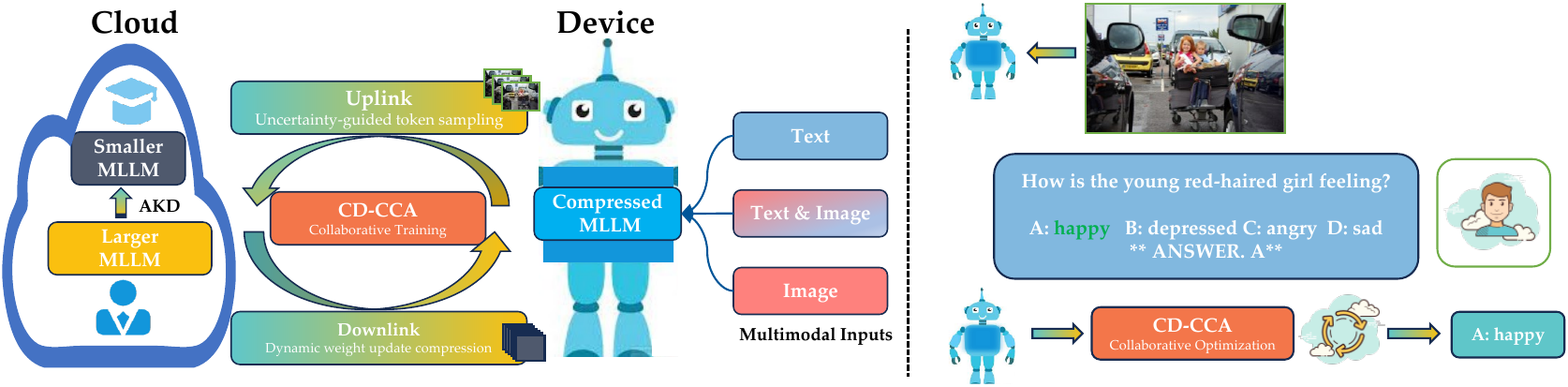} 
\caption{\textbf{Cloud-Device Collaborative Continual Adaptation framework (CD-CCA).} Our CD-CCA, specifically designed for MLLMs, embodies a cloud-device collaborative paradigm. It is adept at receiving various modalities and executing multimodal comprehension tasks. As illustrated on the left side of the figure, our approach facilitates collaborative learning between device and cloud, enabling the update on the device-side deployed MLLM to adapt to dynamically changing scenarios. On the right side, an application instance of CD-CCA is depicted, demonstrating its capability to achieve accurate multimodal comprehension in the face of evolving scenarios at the device.} 
\label{Fig_intro1} 
\end{figure*}
To empower device models in dynamic environments, we propose a Cloud-Device Collaborative Continual Adaptation (CD-CCA) framework for MLLMs (Figure \ref{Fig_intro1}).
Our key insight is harnessing cloud-larger MLLMs to boost the generalization capability of smaller, compressed MLLMs deployed on the device.
In pursuit of augmenting the generalization capabilities of device models without compromising their efficiency, as well as facilitating their dynamic adaptability to ever-changing distributions, we propose a new learning paradigm: Cloud-Device Collaborative Continual Adaptation.
The paradigm has three key components: the device-to-cloud uplink, the cloud-side knowledge update, and the cloud-to-device downlink.

First, in order to enable the MLLM deployed on devices to have the capability of dynamic parameter updating, we design a device-to-cloud uplink for transmitting uncertainty tokens generated on the device side. Specifically, we propose a coarse-to-fine token filtering approach known as the Uncertainty-guided Token Sampling (UTS) strategy to minimize upstream transmission costs. We begin by utilizing sample-level uncertainty to identify and filter out corner case samples from the target distribution data. Subsequently, we adopt token-level uncertainty to perform a secondary filtering process, isolating out-of-distribution tokens. This approach helps alleviate network transmission bandwidth constraints and enhances training efficiency on the cloud server.

Second, on the cloud side, we develop a novel Adapter-based Knowledge Distillation (AKD) method, specially designed for MLLMs. The purpose of AKD is to transfer dark knowledge from the original huge MLLMs to the compressed pocket-size MLLMs. MLLMs typically consist of three main components: a vision encoder, a large language model (LLM) \cite{touvron2023llama}, and a cross-modal transformer, which fuses the high-level vision and language context~\cite{li2023blip2,Alayrac2022Flamingo,zhang2023llama}. Therefore, our approach initially focuses on conducting KD for the learnable query adapter from the cross-modal transformer, enhancing the small MLLMs' vision-to-text alignment capabilities.
Simultaneously, since the LLM occupies the majority of parameters in MLLM, the primary objective for the compressed model is to reduce the LLM's parameter. Consequently, we further conduct KD for learnable language adapters, which are plugged into the LLM, to enhance the student MLLMs' language communication and reasoning abilities.

Furthermore, to account for the varying computational capabilities of edge devices, we employ an adaptive quantization and compression technique for the dynamically updated weight parameters for the device-side MLLMs. These compressed weight parameters are then transmitted to the device through the downlink, narrowing the gap in representation between the device and cloud MLLMs.
We conducted extensive experiments on two cross-domain visual reasoning benchmarks, one from VQA-v2 \cite{antol2015vqa} to A-OKVQA \cite{schwenk2022okvqa} and the other from COCO Captions 2017 \cite{chen2015microsoft} to nocaps \cite{agrawal2019nocaps}. Our proposed framework achieved superior performance compared to previous methods.
Additionally, for the uplink, we maintain the performance while reducing transmission costs to 4.71\% and 20.6\% compared to transferring the entire dataset. As for the downlink, we can deliver the compressed dynamically updated weight parameters with almost negligible transmission cost to the device, resulting in 3.93\% and 2.20\% improvements in domain-shifted VQA tasks and captioning tasks.
Our contributions can be summarized as follows:
\begin{itemize}
    \item We introduce the CD-CCA framework that involves the continuous utilization of cloud-based large MLLMs to enhance the generalization capabilities of smaller, compressed MLLMs on the device.
    \item For the device-to-cloud uplink, we propose UTS strategy, which serves to filter out-of-distribution tokens during data transmission from the device to the cloud.
    \item On the cloud side, we introduce the AKD manner to facilitate the transfer of dark knowledge from the original huge MLLMs to the compressed pocket-size MLLMs.
    \item For the cloud-to-device downlink, we propose a dynamic weight updating compression method that significantly enhances the transmission efficiency of updated weights from cloud to device, which establishes a practical foundation for the application of the Cloud-Device collaborative learning paradigm.
    \item Extensive experiments demonstrate CD-CCA outperforms previous methods, effectively enhancing the continuous domain adaptation capability of device-compressed MLLMs. Moreover, we validate the feasibility of our approach through real-world experiments.
\end{itemize}

\section{Related Work}

\textbf{MLLMs.} 
Recent advancements in LLMs\cite{touvron2023llama,brown2020language} have marked a shift from single to multi-modal capabilities, with MLLMs~\cite{li2023blip2,wang2022ofa,Alayrac2022Flamingo} emerging as a significant development. 
However, this expansion has led to increased model sizes, escalating training costs to prohibitive levels. 
Despite the efforts to minimize the trainable parameters~\cite{houlsby2019parameterefficient}, model deployment on device continues to pose significant challenges, constrained by limited computational power and network bandwidth. In this work, we conceive a new training strategies to replicate the magic of large models in resource-constrained environments.
\\
\textbf{Cloud-Device Collaborative Learning.} 
Previous approaches have attempted to offload the computational workload to the cloud~\cite{chinchali2019network, crankshaw2017clipper, 211251, kang2017neurosurgeon,gan2022clouddevice}, effectively reducing the hardware requirements on devices. However, these methods usually represent a superficial level of cloud-device collaboration. 
Our method introduces the UTS strategy, designed to filter out-of-distribution image tokens from devices to cloud, which significantly reduces the required upstream bandwidth while ensuring that the selected image tokens are rich in semantic information.

Knowledge Distillation (KD) methods have been proposed that perform distillation over intermediate features \cite{huang2022masked, huang2023knowledge}, relation representation \cite{2020FKD, yang2022cross}, attention \cite{yang2022focal, zhang2023avatar}. However, for MLLMs, there is currently no specific knowledge distillation method available to compress them effectively. 


\textbf{Continual Domain Adaptation.} 
Devices are commonly deployed in real-world scenarios where data is continuously evolving. In recent years, several works have been proposed to continually adapt the model to the changing target domain~\cite{wang2021tent,liang2021really,li2020model,liu2021sourcefree}.
Our work proposes a Cloud-Device Collaborative Continual Adaptation framework, enabling the model to adapt to dynamically changing distributions. This approach allows for the simultaneous improvement of the teacher model in cloud and student model on devices.

\label{sec:related}
\begin{figure*}[ht] 
\centering 
\includegraphics[width=0.85\textwidth]{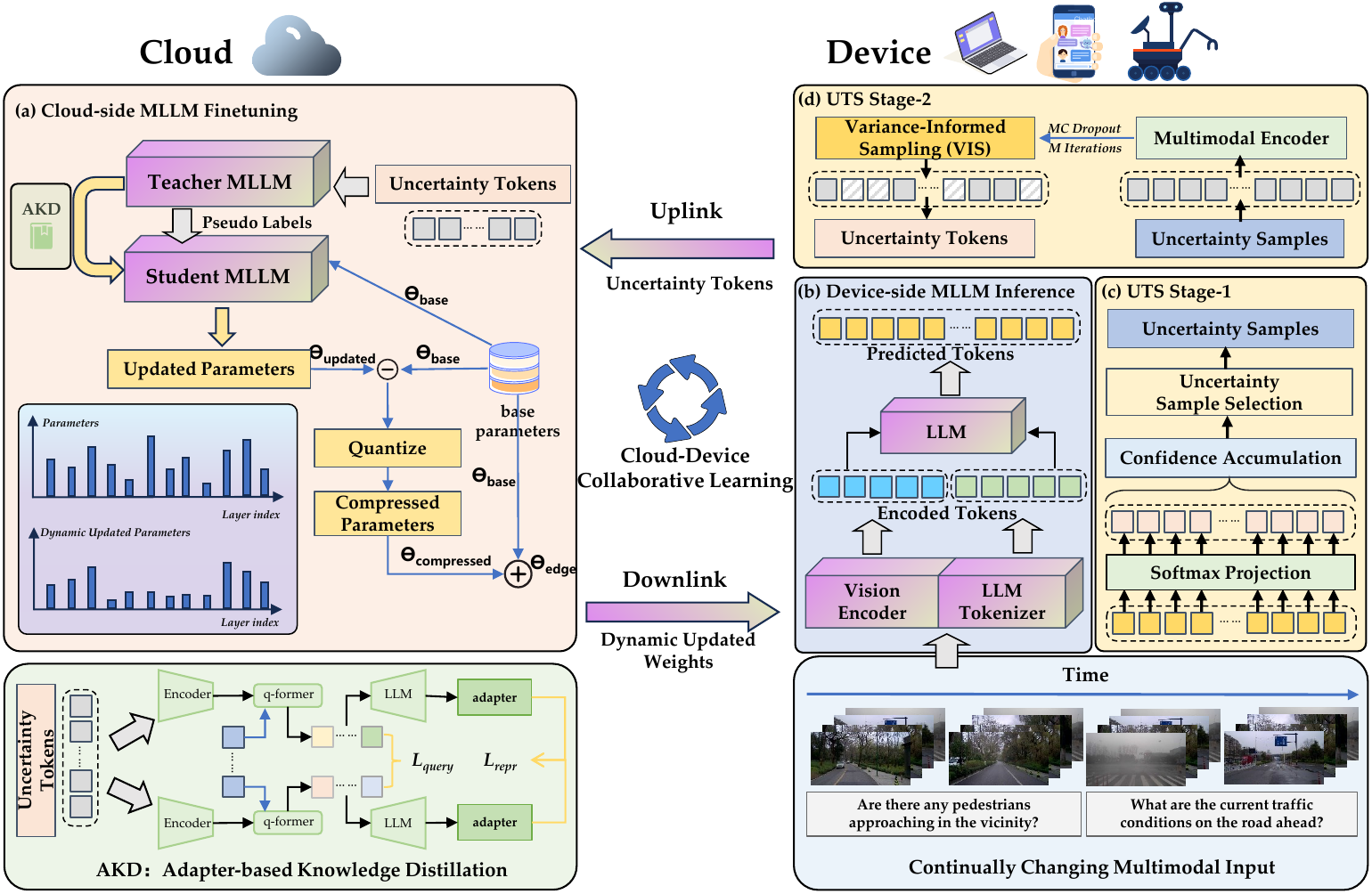} 
\caption{\textbf{The overall pipeline of CD-CCA.} 
(a) Cloud: Upon receiving a token from the device, the Teacher MLLM generates pseudo labels and distills knowledge for the smaller model. (b) Device: Upon receiving an image and a human prompt, it generates the corresponding answer. (c) First stage of Uncertainty-guided Token Sampling (UTS). (d) Second stage of UTS.}
\label{Fig_method} 
\end{figure*}
\section{Approach}
\label{sec:approach}
In this section, we propose CD-CCA to enhance device-deployed MLLMs through efficient cloud-device collaboration. We describe the overall pipeline in Sec. \ref{sec:app_COF}, 
and then introduce the key components in the following subsections. 

\subsection{Overview of CD-CCA Framework}
\label{sec:app_COF}
In the landscape of pervasive computing, edge devices are increasingly tasked with complex multimodal interactions, necessitating models that are not only robust but also adaptive to continual environmental shifts. The CD-CCA framework, shown in Figure \ref{Fig_method}, emerges as a paradigm designed to synergize the computational prowess of cloud resources with the operational nimbleness of edge devices. 
This dynamic adaptability of the CD-CCA framework can be succinctly encapsulated in the following optimization process:
\begin{equation}
\mathcal{M}' = C\left(K\left(U\left(\mathcal{D}, \mathcal{M}_{\text{edge}}\right), \mathcal{M}_{\text{cloud}}^{\text{teacher}}\right), \mathcal{M}_{\text{cloud}}^{\text{student}}\right)
\end{equation}

where $\mathcal{M}'$ signifies the refined model deployed back on the edge device, $\mathcal{D}$ represents the dataset of multimodal instances, $U$ delineates the UTS for uplink efficiency, $K$ depicts the AKD on the cloud, and $C$ denotes the Dynamic Weight update Compression (DWC) for the downlink transmission.

Initially, the framework employs UTS, a novel approach that discernibly filters the influx of multimodal data, earmarking only the most pivotal tokens for cloud-assisted refinement. The selective process is pivotal in distilling the essence of data that demands the cloud's attention, thereby conserving bandwidth and reducing uplink latency. Subsequently, the framework leverages an AKD technique in the cloud, which distills and transfers the rich knowledge from an expansive teacher model to a compact student counterpart. The AKD process is fine-tuned to cater to the specific learning nuances of multimodal data, ensuring that the student model is endowed with enhanced generalization capabilities. Culminating the framework's process is DWC, an innovative strategy that dynamically quantizes and compresses the updated model parameters before transmission via the downlink, significantly alleviating the latency typically associated with updating device-resident models. The DWC ensures that the updated intelligence is delivered promptly, maintaining the real-time responsiveness crucial for device applications. Collectively, these components of the CD-CCA framework constitute a powerful conduit for continual learning, enabling MLLMs to evolve in situ, with a level of acuity and efficiency previously unattainable in device computing paradigms.

\subsection{Uncertainty-guided Token Sampling (UTS)}
\label{sec:app_UTS}
As devices operate within the intrinsic variability of real-world scenarios, there is a crucial need for the continual adaptation of MLLMs that can process data selectively, concentrating computational efforts where they are most needed. To this end, the UTS component of the CD-CCA framework serves as an intelligent filtration mechanism, enabling the discernment and prioritization of multimodal instances for transmission. This is rooted in the understanding that not every instance contributes equally to the model's learning and that some may be more pivotal for adaptation.

In the first stage of UTS, an MLLM with parameters \( \Theta \) deployed on an edge device processes a multimodal instance \( (v_i, t_i) \in \mathcal{D} \), and its predictive uncertainty \( \mathcal{U} \) is evaluated as follows:
\begin{equation}
\label{eq:3_2}
\mathcal{U}(v_i, t_i; \Theta) = -\sum_{j} p(y_{ij} | v_i, t_i; \Theta) \log p(y_{ij} | v_i, t_i; \Theta)
\end{equation}

Eq. \ref{eq:3_2} calculates the entropy of the predicted token probabilities, which serves as a measure of uncertainty for the given instance. Instances with high uncertainty are flagged as candidates for further analysis.

In the subsequent phase, we propose Variance-Informed Sampling (VIS) technique as a refinement step to further sift through the pre-selected instances. VIS applies Monte Carlo dropout to the encoded multimodal input tensors, deriving a variance measure across multiple forward passes to identify which tokens within these instances exhibit significant variability in their representations:
\begin{equation}
\label{eq:3_3}
\sigma^2(v_i, t_i; \Theta) = \frac{1}{M} \sum_{m=1}^{M} \left( \mathcal{F}_m(v_i, t_i; \Theta) - \bar{\mathcal{F}}(v_i, t_i; \Theta) \right)^2
\end{equation}

Here, tokens with a variance \( \sigma^2 \) exceeding a predefined threshold \( \beta \) are retained, ensuring that only the most informative tokens are considered for cloud processing, as shown in Eq. \ref{eq:3_4}:
\begin{equation}
\label{eq:3_4}
\tau(\sigma^2(v_i, t_i; \Theta), \beta) = 
\begin{cases} 
1, & \text{if } \sigma^2(v_i, t_i; \Theta) > \beta \\
0, & \text{otherwise}
\end{cases}
\end{equation}

By implementing this two-stage approach, UTS significantly reduces the volume of data required for uplink transmission, thereby optimizing bandwidth usage and minimizing latency. The VIS, in particular, plays a critical role by ensuring that the model's enhancement is driven by data points that are likely to contribute the most to its learning progress, embodying the essence of targeted and efficient learning within the CD-CCA framework.

\subsection{Adapter-Based Knowledge Distillation (AKD)}
\label{sec:app_AKD}
The AKD strategy hones the capabilities of device-deployed MLLMs by leveraging the computational abundance of cloud resources. In this process, a high-capacity teacher MLLM and a structurally identical student MLLM coexist on the cloud, engaging in a targeted knowledge transfer. This exchange is facilitated by adapters—auxiliary linear layers that introduce minimal parameters to the model while providing pathways for significant updates.

During the AKD phase, we focus on fine-tuning the student model \( \mathcal{M}_{\text{student}} \) to encapsulate the high-level multimodal comprehension exhibited by the teacher model \( \mathcal{M}_{\text{teacher}} \). Specifically, the adapters are employed to fine-tune the query representations and the cross-attention outputs, which are critical for processing and integrating multimodal information. These adapters act as targeted modification modules, aligning the student's latent space with the teacher's refined feature space, effectively compressing the teacher's extensive knowledge into the student's more concise structure.

This fine-grained distillation process is facilitated through adapters that are strategically placed to intercept and transform the query vectors and the attention-mediated multimodal representations. By so doing, the adapters enable a direct knowledge flow from the teacher's rich feature space to the student's corresponding layers, ensuring the retention of critical multimodal insights.

The effectiveness of this adapter-based fine-tuning is measured by a composite loss function, comprising:

\textbf{Query Alignment Loss (\( \mathcal{L}_{query} \))}: Minimizes the difference between the query representations of the student and teacher models, thereby ensuring that the student can generate queries that encapsulate the complexity of the multimodal data as effectively as the teacher. Regularly, $\bm{Q}^{(t)}\in\mathbb{R}^{B\times L\times C}$ and $\bm{Q}^{(s)}\in\mathbb{R}^{B\times L\times C_s}$ denote the feature maps of teacher and student queries respectively, and the Query Alignment imitation can be fulfilled via: 
\begin{equation} \label{eq:mimic}
    \mathcal{L}_\mathrm{query} = \frac{1}{BLC} \left\|\bm{Q}^{(t)} - \phi(\bm{Q}^{(s)})\right\|_2^2 ,
\end{equation}
where $\phi$ is a linear projection layer to adapt $\bm{Q}^{(s)}$ to the same channels as $\bm{Q}^{(t)}$.

\textbf{Representation Alignment Loss (\( \mathcal{L}_{repr} \))}: Aims to synchronize the attention-driven multimodal representations between the student and teacher models, enhancing the student's ability to process and integrate multimodal cues.

\textbf{Cross-Entropy Loss (\( \mathcal{L}_{CE} \))}: Utilizes the teacher model's output on challenging multimodal instances, which have been identified and transmitted via the uplink after UTS, as pseudo-labels. These labels serve to calibrate the student model's parameter updates, enhancing its capacity to address the complexities inherent in multimodal data. The inclusion of UTS-selected instances ensures that the student model focuses its learning on the data points that are most indicative of its current limitations, thereby promoting a more efficient and targeted learning process.

The distillation procedure optimizes a weighted sum of these loss components, carefully calibrated to achieve a harmonious balance between mimicking the teacher's output and maintaining the student's intrinsic characteristics:
\begin{equation}
\label{eq:3_5}
\mathcal{L}_{total} = \lambda_{query} \mathcal{L}_{query} + \lambda_{repr} \mathcal{L}_{repr} + \lambda_{CE} \mathcal{L}_{CE}
\end{equation}

By minimizing \( \mathcal{L}_{total} \), AKD ensures that the student MLLM not only accurately reflects the teacher's adeptness in handling multimodal data but also remains agile and efficient, key for deployment within the resource-constrained environments typical of device computing.

\subsection{Dynamic Weight update Compression (DWC)}
\label{sec:app_DUC}
DWC forms an integral pillar of the CD-CCA framework, addressing the transmission efficiency of model updates from cloud to device. DWC specifically targets the challenge of bandwidth constraints and latency in updating device-deployed MLLM by introducing a quantization-based compression mechanism for model parameters.

DWC operates on the premise that efficient model updates are not solely contingent on the volume of data transmitted but also on the significance of the parameters updated. This leads to the development of a quantization scheme that selectively targets the parameters refined during the AKD phase, optimizing the update payload for transmission efficiency without compromising the model's performance integrity.

The DWC process can be formalized through the following quantization operation:
\begin{equation}
\Theta_{\text{compressed}} = \text{Quantize}(\Theta_{\text{updated}} - \Theta_{\text{base}}, \mathcal{Q})
\end{equation}

Here, \( \Theta_{\text{updated}} \) represents the parameters post-AKD, \( \Theta_{\text{base}} \) denotes the pre-update baseline parameters, and \( \mathcal{Q} \) is the quantization function that adaptively maps parameters to a compact, lower-bit representation. This function is meticulously calibrated to ensure that the most critical updates are preserved, while the overall update size is reduced.

The quantization process strategically applies a higher compression ratio to less impactful parameters, while preserving the fidelity of more significant updates:
\begin{equation}
\Theta_{\text{edge}} = \Theta_{\text{base}} + \Theta_{\text{compressed}}
\end{equation}

The edge device, upon receiving \( \Theta_{\text{compressed}} \), integrates these updates directly into the MLLM. This direct integration circumvents the need for dequantization, as the device MLLM operates effectively within the quantized parameter space, reflecting the nuanced enhancements learned through cloud-based distillation.

DWC thus enables a practical and scalable approach to model updating in device computing environments, where transmission overhead is a critical concern. By facilitating smaller, yet impactful updates, DWC ensures that the device-deployed MLLMs can continually evolve and adapt to new data without the latency typically associated with large-scale model retraining or full-model updates.
\begin{algorithm}[t]
\caption{Collaborative Learning in CD-CCA}
\label{algorithm1}
\hrulefill 
\begin{algorithmic}[1]
\State Initialize edge model \( \mathcal{M}_{\text{edge}} \) with parameters \( \Theta_{\text{edge}} \)
\State Deploy teacher model \( \mathcal{M}_{\text{teacher}} \) and student model \( \mathcal{M}_{\text{student}} \) on cloud
\State Define UTS, AKD, and DWC procedures
\Repeat
    \State Edge performs inference and UTS to identify high-uncertainty instances
    \State Transmit selected instances to cloud
    \State Cloud performs AKD, utilizing \( \mathcal{M}_{\text{teacher}} \) to refine \( \mathcal{M}_{\text{student}} \)
    \State Compress updated parameters \( \Theta_{\text{updated}} \) using DWC to obtain \( \Theta_{\text{compressed}} \)
    \State Transmit \( \Theta_{\text{compressed}} \) back to device
    \State Update \( \mathcal{M}_{\text{edge}} \) with \( \Theta_{\text{compressed}} \)
\Until{convergence or a predefined number of cycles are completed}
\end{algorithmic}
\hrulefill 
\end{algorithm}

\subsection{Collaborative Learning Strategy}
\label{sec:app_COS}
The essence of CD-CCA resides in its Collaborative Learning Strategy, a synergistic approach that harmonizes the model refinement process across cloud and device platforms, shown in Algorithm \ref{algorithm1}. This strategy encapsulates the concerted efforts of edge devices and cloud services to perpetually enhance the MLLMs seamlessly and efficiently.
The optimization pivots on two key fronts: the edge devices perform UTS to identify and forward challenging multimodal instances to the cloud, while the cloud engages in AKD and DWC to refine and compress the parameter updates, respectively. The culmination of this process is the application of compressed updates to the device-side MLLM, ensuring it remains adept and up-to-date with minimal transmission overhead.
The Collaborative Learning Strategy is a testament to the potential of CD-CCA in fostering a dynamic learning environment where edge-deployed MLLMs can thrive. By leveraging the strengths of both cloud and device computing, it stands as a paradigmatic shift towards more intelligent and adaptable multimodal interactions in real-world applications.

\section{Experiments}
\label{sec:exp}

\subsection{Experimental Setups}
\label{sec:exp_setups}
\textbf{Datasets.} To validate the persistent generalization ability of our proposed CD-CCA for multimodal large language model (MLLM) in the scenario of language domain-shifted distribution, we conducted experiments based on two pairs of datasets, VQA-v2 \cite{antol2015vqa}, A-OKVQA \cite{schwenk2022okvqa}.
and COCO Caption 2017 \cite{chen2015microsoft}, Nocaps \cite{agrawal2019nocaps}.\\
\textbf{Evaluation Metrics.} To demonstrate the MLLM's persistent generalization capability under the proposed CD-CCA and other SOTA domain adaptation methods, VQA Accuracy, BLeU-4, and CIDEr scores are uniformly used as the evaluation metrics. In addition, in real-world validations, we further calculate the quantity of transmitted parameters and data size in the uplink and downlink of CD-CCA, as well as the Cloud-Device transfer delay (TD), respectively.
\\
\textbf{Implementation Details.}
In our experiments, we use LLaMA-Adapter \cite{gao2023llama} with LLaMA2-13B \cite{touvron2023llama} as the large teacher MLLM on the cloud, and we employ LLaMA-Adapter \cite{gao2023llama} with LLaMA2-7B \cite{touvron2023llama}  as the small student MLLM (same as the device model). In addition, to further reduce the quantity of device-side model parameters, we reduce the student MLLM's Q-former \cite{li2023blip} hidden layers, from 12 to 6. The above MLLMs are first pre-trained on large-scale image-text pairs: COYO \cite{kakaobrain2022coyo-700m}, LAION \cite{schuhmann2022laion}, CC3M \cite{sharma2018conceptual}, CC12M \cite{changpinyo2021conceptual}, SBU \cite{ordonez2011im2text}. Then, they are further tuned with 52K single-turn instruction data from GPT4-LLM \cite{peng2023instruction} and 567K captioning data from COCO Caption \cite{chen2015microsoft}. For both cloud and device models, all the parameters in LLaMA normalization layers, linear layer bias, LoRA \cite{hu2021lora}, and query tokens in Q-Former \cite{li2023blip} are set to be updated during finetuning with the remaining parameters kept frozen. In the specific experiments, we further finetuned the MLLMs on the corresponding datasets elaborated before.

\begin{table*}[h]
\centering
\caption{\textbf{Persistent generalization capability on VQAv2-to-AOKVQA.} Visual question-answering results are evaluated on the VQAv2-to-AOKVQA online continual test-time adaptation task. MC and DA are VQA accuracy ($\%$) calculated following \cite{schwenk2022okvqa} under different conditions (multiple choices and direct answers). Gain ($\%$) refers to the accuracy improvement compared with the source-only method.}
\label{exp:vqa_tab1}
\tabcolsep=0.1cm
\begin{tabular}{l|cccccc|cccc}
\hline
Time & \multicolumn{10}{|c}{ \textit{t} \tikz[baseline=-0.5ex]{\draw[->, line width=0.1mm] (0,0) -- (\linewidth*0.6,0);}}    \\
\hline
Round & \multicolumn{2}{|c}{$\mathrm{1}_{\textit{st}}$} & \multicolumn{2}{|c}{$\mathrm{2}_{\textit{nd}}$} & \multicolumn{2}{|c}{$\mathrm{3}_{\textit{rd}}$} & \multicolumn{4}{|c}{All} \\
\hline
Condition & MC & DA & MC & DA & MC & DA & $\mathrm{Mean}_{\textit{MC}}$ & $\mathrm{Mean}_{\textit{DA}}$ & $\mathrm{Gain}_{\textit{MC}}$ & $\mathrm{Gain}_{\textit{DA}}$ \\
\hline
Source-only \cite{gao2023llama}     & 47.95 &  45.55 & 47.95 & 45.55  & 47.95 & 45.55 & 47.95 & 45.55 & /     & / \\
TENT-continual \cite{wang2020tent}  & 47.42 &  45.17 & 48.12 & 45.52  & 47.34 & 44.86 & 47.63 & 45.18 & -0.32 & -0.37 \\
CoTTA \cite{wang2022continual}      &  47.77    &  45.02     & 47.77     &  45.30     & 48.30     & 45.02     & 47.95     & 45.11     & +0.00    & -0.44 \\
\hline
PKD \cite{cao2022pkd}    &  48.21   &  45.05     & 48.73     &  45.24     & 47.77     & 45.24     & 48.23     & 45.18     & +0.28     & -0.37 \\
ChannelWiseDivergence \cite{shu2021channel}    &  48.03    &  44.78     &  48.21     &  44.65    & 48.47     & 44.93     & 48.24     & 44.79     & +0.29     & -0.76 \\
\hline
\rowcolor{green!5}
Ours (CD-CCA)  & \textbf{50.65} &  \textbf{48.80} & \textbf{51.79} &  \textbf{48.37} & \textbf{53.19} & \textbf{49.05} & \textbf{51.88} & \textbf{48.74} & \textbf{+3.93}  & \textbf{+3.19} \\
\hline
\end{tabular}
\end{table*}











\begin{table*}[h]
\centering
\caption{\textbf{Persistent generalization capability on COCO-to-nocaps.} Image captioning results are evaluated on the COCO-to-nocaps online continual test-time adaptation task. BLeU@4, CIDEr scores ($\%$) are calculated following \cite{chen2015microsoft} under different conditions (in-domain, near-domain, out-domain, etc.). Gain ($\%$) refers to the improvement compared with the source-only method.}
\label{exp:caption_tab2}
\tabcolsep=0.1cm
\begin{tabular}{l|cc|cc|cc|cc|cc}
\hline
Condition & \multicolumn{2}{|c}{In-domain} & \multicolumn{2}{|c}{Near-domain} & \multicolumn{2}{|c}{Out-domain} & \multicolumn{2}{|c}{All} & \multicolumn{2}{|c}{Gain} \\
\hline
Score & BLeU & CIDEr & BLeU & CIDEr & BLeU & CIDEr & BLeU & CIDEr & BLeU & CIDEr \\
\hline
Source-only \cite{gao2023llama}     & 39.55 &  72.33 & 39.72 & 77.32 & 31.20 & 76.95 & 36.82 & 75.53 & / & /\\
TENT-continual \cite{wang2020tent}  & 39.92    &  71.81     & 39.60     & 74.49     & 30.28     & 72.69     & 36.60 & 73.00 & -0.22 & -2.53\\
CoTTA \cite{wang2022continual}      & 40.12    &  73.87     & 40.08    &76.52     &30.04    & 74.19     & 36.74 & 74.86 & -0.08 & -1.34\\
\hline
PKD \cite{cao2022pkd}    &  39.43    &  73.67    & 39.46     & 76.33     & 31.12     & 76.88     & 36.67 & 75.63 & -0.15 & +0.10\\
ChannelWiseDivergence \cite{shu2021channel}    &  39.03    & 73.82     & 39.10     & 75.87     &  30.15     &75.77     & 36.18 & 75.15 & -0.64 & -0.38\\
\hline
\rowcolor{green!5}
Ours (CD-CCA)  & \textbf{41.34} & \textbf{74.47}  & \textbf{40.67} & \textbf{77.78} & \textbf{33.04} & \textbf{80.93} & \textbf{38.35} & \textbf{77.73} & \textbf{+1.53} & \textbf{+2.20}\\
\hline
\end{tabular}
\end{table*}


\subsection{Comparison Analysis}
\label{sec:exp_comparison}
In this subsection, we conduct comparison experiments between our CD-CCA and the existing SOTA domain adaptation methods \cite{wang2020tent, wang2022continual,cao2022pkd,shu2021channel}.Tent\cite{wang2020tent} updates the trainable parameters in the Batchnorm layer to adapt to the test data by minimizing entropy. Cotta\cite{wang2022continual} employs weight-averaged and augmentation-averaged predictions to reduce the accumulation of errors in pseudo-labeling and utilizes stochastically restore to prevent the issue of catastrophic forgetting.  PKD\cite{cao2022pkd}  utilizes feature imitation based on the Pearson Correlation Coefficient, relaxing constraints on the magnitude of the features while focusing on the relationship information from the teacher. ChannelWiseDivergence\cite{shu2021channel}  normalizes the activation maps of each channel, yielding soft probability maps for the two networks, and minimizes the Kullback-Leibler divergence between the channel probability maps. All the experiments are carried out using LLaMA-Adapter \cite{gao2023llama} as the underlying MLLM. 
\begin{figure}[b]
    \centering
    \includegraphics[width=0.85\linewidth]{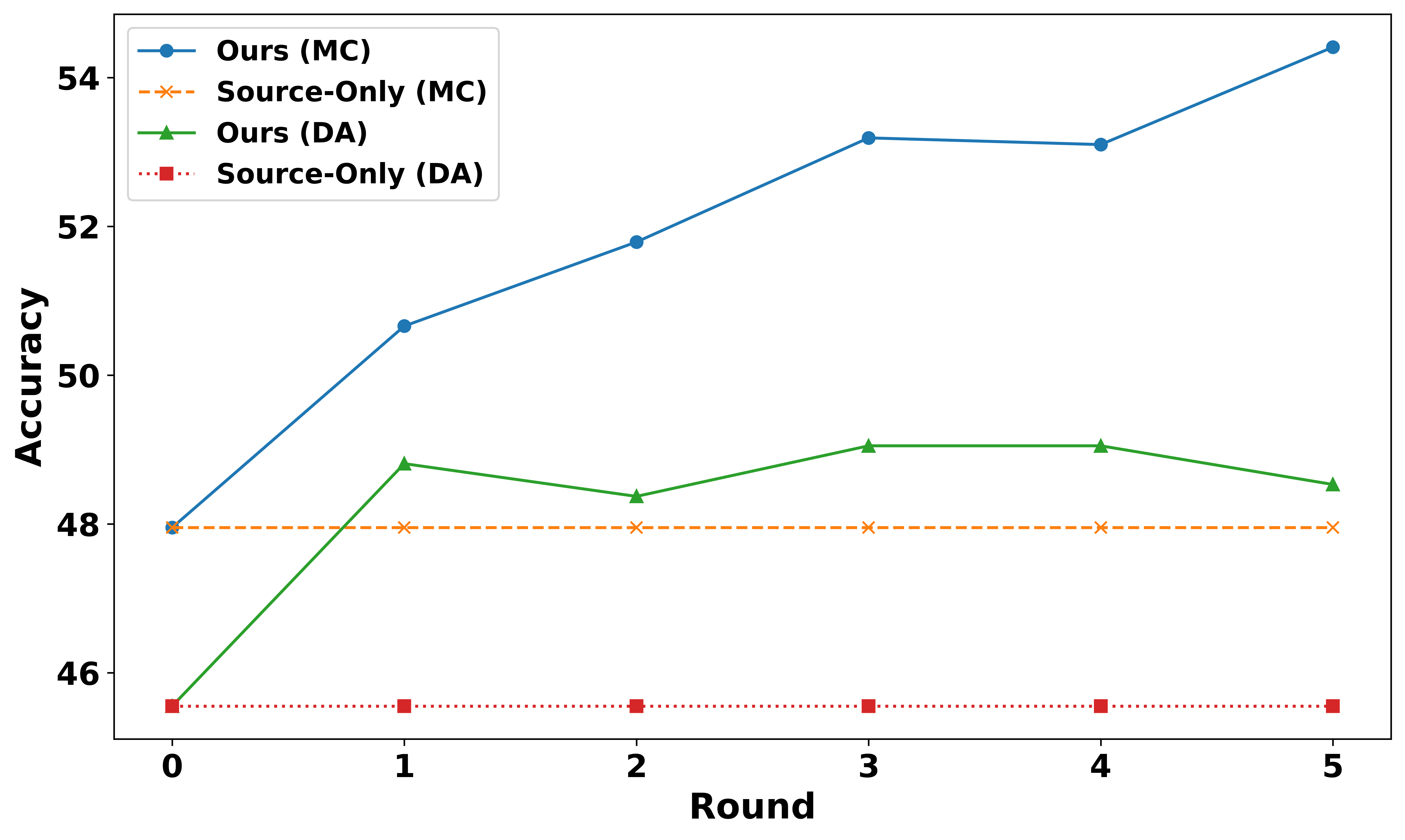}
    \caption{Comparative analysis of CD-CCA and source-only method. The MC and DA accuracy are evaluated over five rounds.}
    \label{fig:combined-accuracy}
\end{figure}

First, to verify the persistent generalization ability of our proposed CD-CCA under the condition of language domain-shifted distribution, we use the VQAv2-to-AOKVQA datasets for evaluation. Specifically, we adopt VQA-v2 \cite{antol2015vqa} to finetune the pre-trained MLLM, LLaMA-Adapter (7B \& 13B). Then, the VQA accuracy results on A-OKVQA \cite{schwenk2022okvqa} under different conditions (multiple choices (MC) \& direct answers (DA)) are evaluated and recorded in Table \ref{exp:vqa_tab1} and Figure \ref{fig:combined-accuracy}.
\begin{figure}[b]
    \centering
    \includegraphics[width=\linewidth]{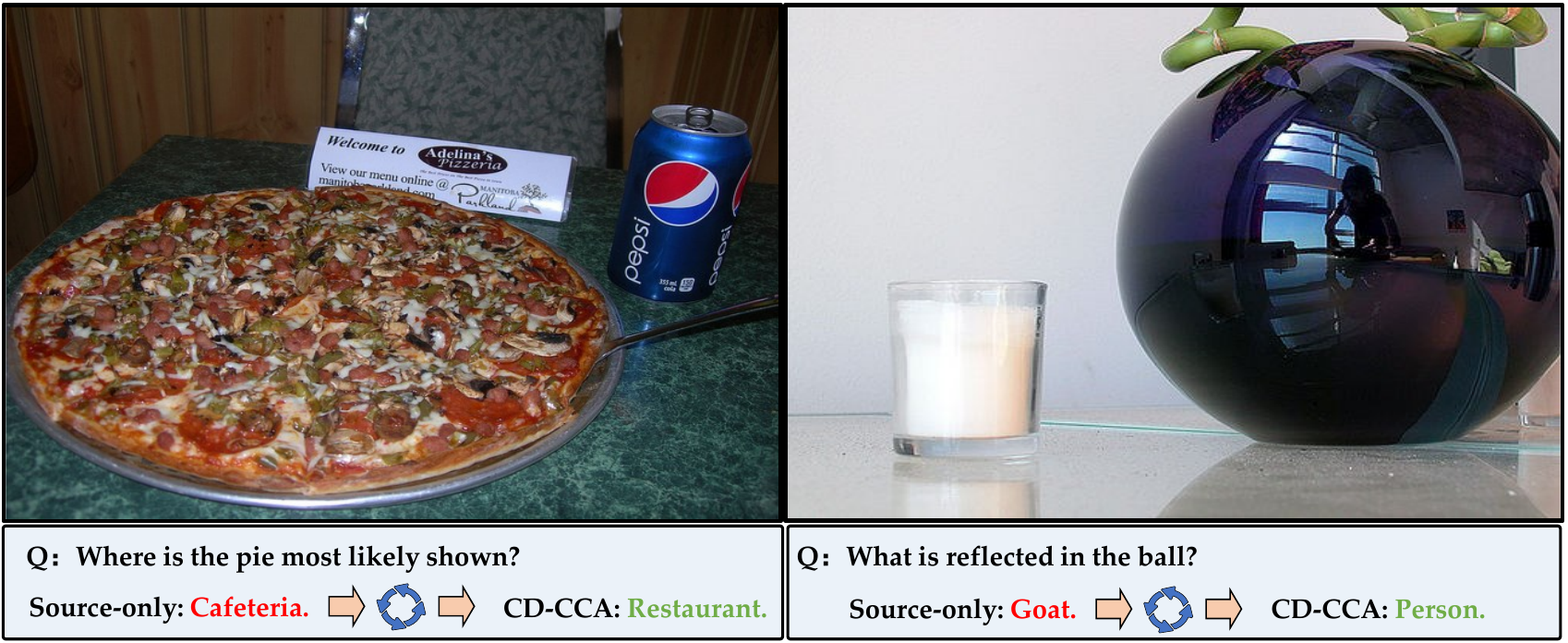}
    \caption{\textbf{Visual results of CD-CCA.} The figure demonstrates the improvement in visual reasoning of device-deployed MLLM facilitated by CD-CCA. 'Source-only' refers to MLLM deployed on the device side without undergoing Cloud-Device Learning.}
    \label{fig:visible}
\end{figure}

In the VQA task, our CD-CCA framework of 1-round scenario has already surpassed the highest accuracy of the comparative models both in MC and DA questions. Notably, we observe that previous methods sometimes lead to performance deterioration. We attribute this to the fact that most of the previous methods were not specifically designed for MLLM, as the model parameter size increases, methods like CoTTA and Tent tend to exhibit a decrease in performance. In contrast, our approach is specifically designed for MLLM, as shown in Table \ref{exp:caption_tab2}, our accuracy is higher by 3.64\% (MC) and 3.19\% (DA) compared to the best-performing comparative model on average. This strongly demonstrates that our framework can maintain a high level of accuracy when faced with constantly changing data distributions. Figure \ref{fig:visible} visually illustrates the experimental results on multimodal comprehension of our proposed framework.

Second, we use the COCO-to-nocaps datasets for further evaluation. We finetune the pre-trained LLaMA-Adapter (7B \& 13B) on the COCO Captions 2017 dataset \cite{chen2015microsoft}. Then, the visual caption results (BLeU@4, CIDEr) on nocaps  \cite{agrawal2019nocaps} are evaluated and recorded in Table \ref{exp:caption_tab2}. Based on the overlap of the training-test image categories, following reference \cite{agrawal2019nocaps}, the test images are categorized into in-domain, near-domain, and out-domain cases. 

In the image caption task, our framework significantly outperforms the best comparative methods in all cases. In the In-domain and Near-domain tasks, our framework surpasses the best comparative method by 1.22\% and 0.59\% (BLeU), 0.6\%, and 0.46\% (CIDEr) respectively. In the out-domain task, our CD-CCA's superiority is even more pronounced, with 1.84\% (BLeU) and 3.98 \%(CIDEr). This reflects the strong generalization ability of our CD-CCA framework, which can effectively help the model extract intrinsic knowledge from images and understand them when transferring to new tasks. Moreover, the experimental results in Table \ref{exp:caption_tab2} reaffirm that previous methods sometimes do not apply to MLLM, while our CD-CCA consistently improves performance. This further reflects the effectiveness of our method specifically designed for MLLM.

\begin{table}[h]
\centering
\caption{\textbf{Ablation studies.} We conduct experiments on VQAv2-to-AOKVQA. PL refers to Pseudo Labels. UTS-1 and UTS-2 represent the first and second stage in UTS, respectively.}
\label{exp:ablation_tab3}
\resizebox{\linewidth}{!}{
\tabcolsep=0.1cm
\begin{tabular}{cccc|cccc}
\hline
PL & AKD        & UTS-1      & UTS-2       & MC  &  DA &  $\mathrm{Gain}_{\textit{MC}}$ & $\mathrm{Gain}_{\textit{DA}}$ \\
\hline
             &            &                   &                    &  47.95   &  45.55   &     /      &    /     \\
\checkmark   &            &                   &                    &  50.48   &  48.89   &   2.53     &    3.34  \\
\checkmark   & \checkmark &                   &                    &  50.39   &  \textbf{49.05}   &   2.44     &    \textbf{3.50}  \\
\checkmark   & \checkmark &   \checkmark      &                    &  50.82   &  48.93   &   2.87     &    3.38  \\
\checkmark   & \checkmark &   \checkmark      &    \checkmark      &  \textbf{53.19}   &  \textbf{49.05}   &   \textbf{5.24}     &    \textbf{3.50}  \\
\hline
\end{tabular}
}
\end{table}

\begin{table}[h]
\centering
\caption{\textbf{Performance (MC/DA) comparison in UTS (Token-Level)}. We report VQA score (MC/DA) using different token sampling strategies. Optimal performance is obtained using the uncertainty token guided sampling (UTS) strategy in Cloud-Device collaboration framework.}
\label{exp:UTS_tab4}
\tabcolsep=0.15cm
\begin{tabular}{lcccccc}
\hline
             &   \multicolumn{2}{c}{25\%}     &      \multicolumn{2}{c}{50\%}  &         \multicolumn{2}{c}{75\%}   \\
\cline{2-7}
             &    MC      &      DA           &    MC      &      DA           &    MC      &      DA               \\
\hline
Random       &      50.74       &      49.46            &    50.13     &       48.40           &       50.91    &       49.09               \\
UTS          &    52.49       &      48.92            &     53.19      &       49.05           &       53.19    &       48.96               \\
Gain         &   +1.75       &      -0.54            &     +3.06      &      +0.65           &       +2.28    &       -0.13               \\
\hline
\end{tabular}
\end{table}

\subsection{Ablation Studies}
\label{sec:exp_ablation}
In this section, we meticulously dissected the proposed CD-CCA framework and its performance across various test scenarios. To gain granular insights into the individual contributions of various components to the framework's efficacy, we systematically dismantle key components. 

\textbf{Effectiveness of UTS strategy.}
Our UTS strategy effectively reduces transmission costs while maintaining performance, as shown in Table \ref{exp:real_tab5}, we achieve the same performance with only 0.21\% in transmission data volume and 0.20\% in transfer latency compared to transmitting the entire dataset. To further validate the effectiveness of UTS, we explore VQA results at different mask ratios, as shown in Table \ref{exp:UTS_tab4}. The model performs best when the mask ratio is set at 50\%. Specifically, we achieved a notable increase of 3.06\% in the accuracy of MC and a 0.65\% increase in the accuracy of DA. Furthermore, we also investigate the effectiveness of each stage in UTS, as shown in Table \ref{exp:ablation_tab3}, and the results indicate that each stage of UTS contributes to improving the performance of the model. When both stages are used in conjunction, there is a significant improvement of 5.24\% and 3.50\% in the MC and DA problems. 

\textbf{Effectiveness of Cloud-device joint optimization with AKD.}
our proposed AKD strategy utilizes adapters for targeted knowledge transfer between teacher and student models, enhancing the generalization ability of the student model. As shown in Table \ref{exp:ablation_tab3}, compared to the pure pseudo-labeling method, AKD improves performance by 2.53\% (MC) and 3.34\% (DA) in VQA tasks, while combining AKD with other modules further enhances performance steadily. The model parameters obtained after AKD are further quantitatively compressed through the DWC method. 

\textbf{Effectiveness of DWC.}
The DWC strategy in the cloud aims to quantitatively compress model parameters, ensuring that only the most effective parameters are updated on the device. This alleviates the performance burden on the device and effectively reduces the amount of data transmitted to the device during downlink. As shown in Table \ref{exp:real_tab5}, compared to no processing, our approach significantly reduces the weight parameter quantity, data quantity, and transmission latency of the model transmitted to the device, by 99.98\%, 99.99\%, and 99.98\% respectively. This effectively guarantees real-time updates of device parameters.
\begin{table}[h]
\centering
\caption{\textbf{Validation of transmission parameters in real machine}. We report a quantitative analysis of bidirectional transmission parameters size (P), transmission data volume (D), and transfer latency (TL) in a real-world robot system. Uplink parameters are calculated with a five-frame input.}
\label{exp:real_tab5}
\tabcolsep=0.25cm
\begin{tabular}{lcccc}
\hline
             &   \textit{P}    &  \textit{D}        &   \textit{TL}         \\
\hline
Uplink       &      /                       &   31.10 MB         &   0.498s              \\ 
Uplink-UTS   &      /                       &   65.54 KB         &   0.001s              \\ 
Downlink     &      7.78B                   &   14.48 GB         &   65.490s             \\
Downlink-DWC &      1.65M                   &   0.791 MB         &   0.013s              \\ 
\hline

\end{tabular}
\end{table}






\subsection{Real-world Validations}
We utilized Gigabit Ethernet as the actual network environment with a theoretical peak of 1000Mbps, adhering to the 802.11ac (Wi-Fi 5) standard. We employed the Realsense D435i as the image capture device on the device, collecting images at a resolution of 1920$\times$1080. The effectiveness of our CD-CCA was further validated through experiments on a real machine, as shown in Table \ref{exp:real_tab5}, which includes the bidirectional transmission parameters size (P), transmission data volume (D), and transfer latency (TL). 
\label{sec:exp_real_world}
\section{Conclusion}
\label{sec:conclusion}
We propose CD-CCA to empower device models in dynamic environments, including components of uncertainty-guided token sampling, adapter-based KD, and dynamic weight update compression.
Experimental results in the open-world scenario demonstrate performance improvements of 3.93\% in domain-shifted VQA tasks and 2.20\% in captioning tasks.

{
    \small
    \bibliographystyle{unsrtnat}
    \bibliography{main}
}
\clearpage
\setcounter{page}{1}
\maketitlesupplementary
\appendix

\section{Overview}
\label{sec:rationale}
To enhance the completeness of our experiments, additional analyses on the generalization capabilities of the MLLM under various domain shifts are included in the supplementary material. These analyses are conducted from two perspectives: image-domain and text-domain, specifically illustrating the enhancement in the multimodal understanding capabilities of our approach when dealing with domain-shift scenarios. The following aspects are included in our supplementary material.

\begin{itemize}
    \item Supplementary Experimental Analysis
    
    \noindent\hspace*{0.2em}- Generalization capabilities in text-domain shifts

    \noindent\hspace*{0.2em}- Generalization capabilities in image-domain shifts

    \noindent\hspace*{0.2em}- Generalization capabilities in multimodal-domain shifts
    
    \item Additional Visualization Results
    
    \noindent\hspace*{0.2em}- Visualization results in text-domain shifts

    \noindent\hspace*{0.2em}- Visualization results in image-domain shifts
    
    \noindent\hspace*{0.2em}- Visualization results in multimodal-domain shifts
    
    \item Expanded Related Work
    
    \noindent\hspace*{0.2em}- Expanded related work in MLLMs

    \noindent\hspace*{0.2em}- Expanded related work in \textbf Cloud-Device Collaborative Learning

    \noindent\hspace*{0.2em}- Expanded related work in \textbf Continual Domain Adaptation.
    
    \item Demo Video and Dataset

\end{itemize}
\section{Supplementary Experimental Analysis}
\subsection{Generalization capabilities in text-domain shifts}
To validate the enhancement of the generalization performance of MLLMs deployed on the device-side by our method, we employed the VQAv2-to-AOKVQA to simulate text-domain distribution shifts in the real world. Initially, both the pocket-size MLLM on device-side and the teacher MLLM in the cloud were fine-tuned using the VQAv2 dataset. Subsequently, we utilized the AOKVQA dataset for testing, aiming to mimic scenarios in the open world where the image-domain remains largely unchanged while the text-domain input varies. The results, as presented in the main text, distinctly show that our approach demonstrates a notable advantage in enhancing the continual generalization capability of multimodal large models amidst changing text-domain inputs, compared to other recent domain-adaptation and knowledge distillation methods. 


\begin{table*}[h]
\centering
\caption{\textbf{Persistent generalization capability on VQAv2-IDS (image-domain shift).} During the training phase, we fine-tuned the MLLM using the VQAv2 dataset. For testing, we employed a newly constructed dataset, VQAv2-IDS, which introduces random variations of rain, snow, fog, and lighting adjustments to the images while retaining the original VQAv2 question information. The VQAv2-IDS dataset represents image-domain alterations designed to simulate various environmental changes within the image domain in an open-world setting. DA is VQA accuracy ($\%$) calculated following \cite{schwenk2022okvqa} under direct answers. Gain ($\%$) refers to the accuracy improvement compared with the source-only method.}
\label{exp:supply_image_domain_tab}
\tabcolsep=0.5cm
\begin{tabular}{l|ccc|cc}
\hline
Time & \multicolumn{5}{|c}{ \textit{t} \tikz[baseline=-0.5ex]{\draw[->, line width=0.1mm] (0,0) -- (\linewidth*0.3,0);}}    \\
\hline
Round & $\mathrm{1}_{\textit{st}}$ & $\mathrm{2}_{\textit{nd}}$ & $\mathrm{3}_{\textit{rd}}$ & $\mathrm{Mean}_{\textit{DA}}$ & $\mathrm{Gain}_{\textit{DA}}$\\
\hline
Source-only \cite{gao2023llama}     & 35.47 &  35.47 &  35.47 & 35.47 &   /     \\
TENT-continual \cite{wang2020tent}  & 36.04 &  36.20 &  35.86 & 36.03 & +0.56   \\
CoTTA \cite{wang2022continual}      & 35.52 &  35.94 &  35.43 & 35.63 & +0.16   \\
\hline
PKD \cite{cao2022pkd}    & 38.06 & 38.09 & 38.05 & 38.07 &  +2.6  \\
ChannelWiseDivergence \cite{shu2021channel}    & 38.40 & 38.07 & 38.60 & 38.36 & +2.89  \\
\hline
\rowcolor{green!5}
Ours (CD-CCA)  & \textbf{41.41}  & \textbf{41.49} & \textbf{41.64} & \textbf{41.51} & \textbf{+6.04}\\
\hline
\end{tabular}
\end{table*}

\subsection{Generalization capabilities in image-domain shifts}
To assess the enhancement of generalization performance in device-side deployed MLLMs by our method, we utilized the VQAv2-IDS to simulate image-domain distribution shifts in the real world. Initially, both the device-side pocket-size MLLM and the cloud-based teacher MLLM were fine-tuned using the VQAv2 training dataset. We then constructed the VQAv2-IDS dataset to emulate scenarios in the open world where the text-domain remains constant while the image-domain undergoes significant distribution shifts. Specifically, we modified 2,000 of the most difficult instances in VQAv2 test images by randomly introducing natural elements like rain, snow, and fog, and adjusting lighting conditions, creating the VQAv2-image-domain-shift (VQAv2-IDS) test image dataset. Subsequently, without altering the text-domain inputs of the VQAv2 test data, we replaced the image-domain with VQAv2-IDS for comparison against recent state-of-the-art (SOTA) domain-adaptation and knowledge distillation methods, as shown in Table \ref{exp:supply_image_domain_tab}. It is evident that the source-only MLLM deployed on device, constrained by its parameter size, exhibits weaker generalization capabilities for inputs with significant image domain shifts. In contrast, our method rapidly improves the performance of device-deployed MLLMs, achieving sustained generalization.

\subsection{Generalization capabilities in multimodal-domain shifts}



\begin{table*}[h]
\centering
\caption{\textbf{Persistent generalization capability on VQAV2-to-AOKVQA-IDS (multimodal-domain shift).} During the training phase, we fine-tuned the MLLM using the VQAv2 dataset. For testing, we employed a newly constructed dataset, AOKVQA-IDS, which introduces random variations of rain, snow, fog, and lighting adjustments to the images while retaining the AOKVQA's question information. The VQAv2-to-AOKVQA-IDS dataset represents multimodal-domain (image \& text) alterations designed to simulate various environmental changes within both the image and text domain in an open-world setting. MC and DA are VQA accuracy ($\%$) calculated following \cite{schwenk2022okvqa} under different conditions (multiple choices and direct answers). Gain ($\%$) refers to the accuracy improvement compared with the source-only method.}
\label{exp:supply_multimodal_domain_tab}
\tabcolsep=0.2cm
\begin{tabular}{l|cccccc|cccc}
\hline
Time & \multicolumn{10}{|c}{ \textit{t} \tikz[baseline=-0.5ex]{\draw[->, line width=0.1mm] (0,0) -- (\linewidth*0.6,0);}}    \\
\hline
Round & \multicolumn{2}{|c}{$\mathrm{1}_{\textit{st}}$} & \multicolumn{2}{|c}{$\mathrm{2}_{\textit{nd}}$} & \multicolumn{2}{|c}{$\mathrm{3}_{\textit{rd}}$} & \multicolumn{4}{|c}{All} \\
\hline
Condition & MC & DA & MC & DA & MC & DA & $\mathrm{Mean}_{\textit{MC}}$ & $\mathrm{Mean}_{\textit{DA}}$ & $\mathrm{Gain}_{\textit{MC}}$ & $\mathrm{Gain}_{\textit{DA}}$ \\
\hline
Source-only \cite{gao2023llama}     & 46.55 &  43.60 & 46.55 & 43.60  & 46.55 & 43.60 & 46.55 & 43.60 & /     &    /   \\
TENT-continual \cite{wang2020tent}  & 46.72 &  43.75 & 45.50 & 43.75  & 47.24 & 43.06 & 46.48 & 43.52 & -0.07 &  -0.08 \\
CoTTA \cite{wang2022continual}      & 46.98 &  43.62 & 47.42 & 43.53  & 46.81 & 44.03 & 47.07 & 43.72 & +0.52 &  +0.12 \\
\hline
PKD \cite{cao2022pkd}    &  48.64   &  46.38     & 48.38     &  \textbf{46.79}     & 49.25   & 46.69    & 48.76    & 46.62     & +2.21     & +3.02 \\
ChannelWiseDivergence \cite{shu2021channel}    &  48.55    &  46.35    &  49.17     &  46.60    & 49.43     & 46.13     & 49.05     & 46.35     & +2.50    & +2.75 \\
\hline
\rowcolor{green!5}
Ours (CD-CCA)  & \textbf{50.22} &  \textbf{46.88} & \textbf{51.27} &  46.60 & \textbf{51.52} & \textbf{46.94} & \textbf{51.00} & \textbf{46.81} & \textbf{+4.45}  & \textbf{+3.21} \\
\hline
\end{tabular}
\end{table*}

To validate the enhanced generalization performance of our method for MLLMs deployed on the device-side, we utilized the VQAv2-to-AOKVQA-IDS dataset to simulate multimodal-domain distribution shifts in the real world. Initially, both the pocket-size MLLM at the device-side and the cloud-based teacher MLLM were fine-tuned using the VQAv2 training dataset. In the testing phase, we employed images with image-domain shifts and used AOKVQA's input text as the input for the other modality. This approach helped establish a multimodal domain gap with the training data, simulating multimodal domain shifts in the open world. The VQAv2-to-AOKVQA-IDS dataset encompasses images with significant image-domain shifts and texts with notable text-domain shifts. Based on this, we conducted a series of comparative experiments, the results of which are shown in Table \ref{exp:supply_multimodal_domain_tab}. Our method effectively improves the generalization performance of device-side deployed MLLMs in handling multimodal-domain shifts. For scenarios difficult to comprehend by source-only models, significant performance improvements were achieved after a single round of Cloud-Device Collaborative Learning.

With this, we have completed experimental analyses across three different types of domain shifts. Additionally, our experiments on the COCO-to-nocaps dataset, presented in the main text, further validate our CD-CCA framework's sustained generalization capabilities for MLLMs from the perspective of category-domain shift.

\begin{figure*}[ht] 
\centering 
\includegraphics[width=1.00\textwidth]{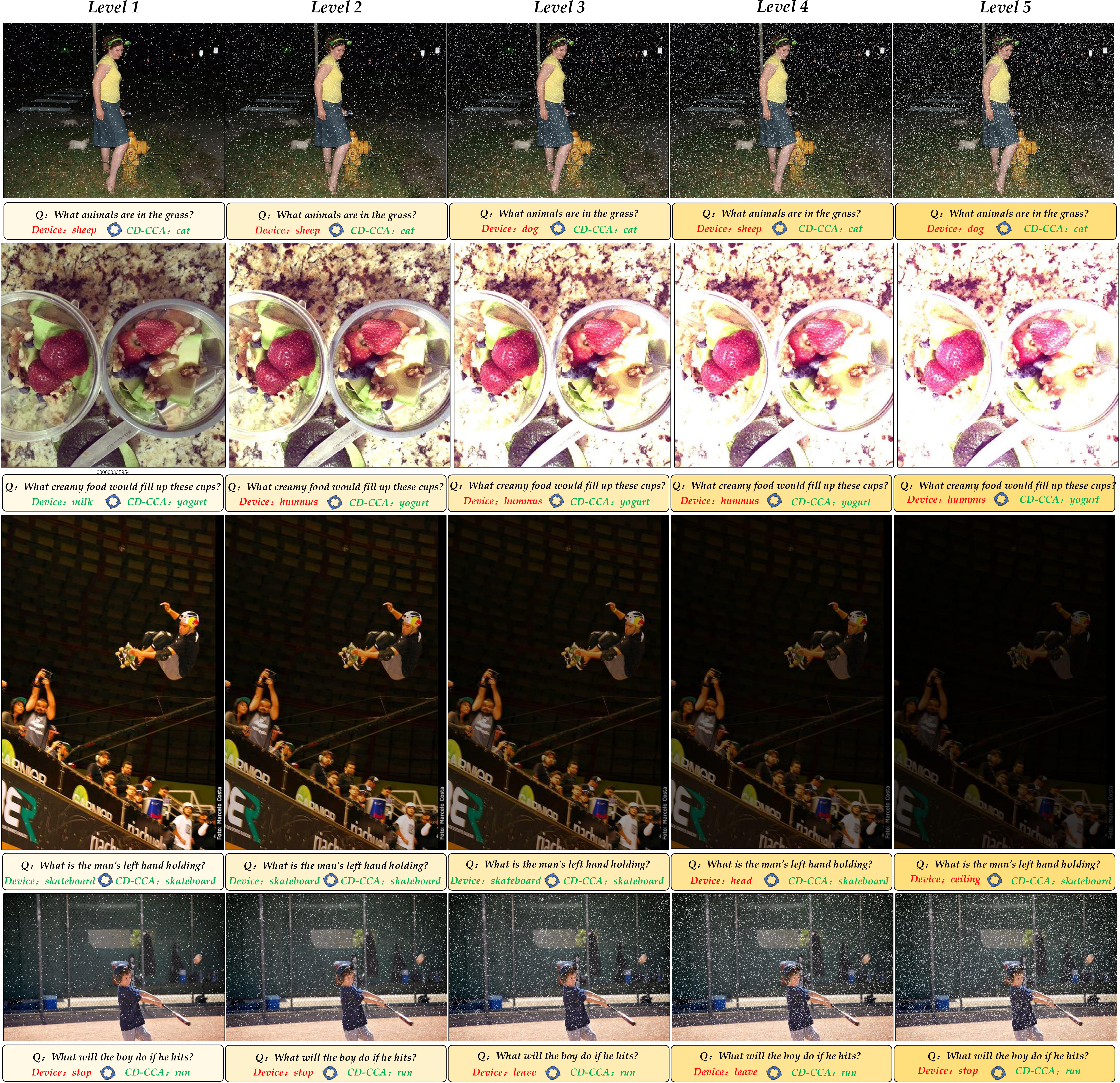} 
\caption{\textbf{Visualization of Experimental Results under Different Domain Shifts.} 
We artificially introduced uncertainty elements (rain, snow, fog, brightness, etc.) into the multimodal inputs to simulate the continuously changing natural environments. The intensity of added uncertainty gradually increases from level 1 to level 5. The resulting figures illustrate the performance of the device-side deployed MLLM in visual question answering tasks, as well as the improved outcomes following the CD-CCA optimization of the device-side MLLM.}
\label{Fig_supply} 
\end{figure*}
\section{Additional Visualization results}
To visually demonstrate the enhanced continual generalization ability of device-side pocket-size MLLMs in handling domain shifts, facilitated by our proposed CD-CCA, we present the following experimental results in a visual format. As shown in Figure \ref{Fig_supply}, we depict the comprehension capabilities of the device-side MLLM under various environments, both before and after the application of CD-CCA. It is clearly observable that, as domain shifts intensify, the generalization ability of the source-only MLLM deployed on the device-side progressively decreases. However, following efficient collaborative learning through CD-CCA, a notable improvement in its generalization capability is achieved.

\section{Expanded Related Work}
\textbf{MLLMs.} Current MLLMs extend beyond linguistic processing by expanding the scale of data and model architecture, enabling real-world perception and addressing limitations in tasks like image captioning \cite{chen2015microsoft} and visual question answering \cite{antol2015vqa}. Due to constraints in model size and training costs, some scholars attempted to predominantly utilize frozen LLM backbones, focusing exclusively on training visual components, or adopting more streamlined and efficient training strategies such as parameter-efficient fine-tuning\cite{houlsby2019parameterefficient} rather than training from scratch. Considering the limitations imposed by computational power and network bandwidth for model deployment on devices, merely reducing the number of model-trainable parameters is insufficient. Therefore, we propose the CD-CCA framework as a solution. 

\textbf{Cloud-Device Collaborative Learning.} 
Merely offloading the computational workload to the cloud without considering the collaboration between the cloud and the device, although alleviating the computational limitations of the device, has minimal impact on enhancing the device's ability to handle complex model processing tasks.
Some work has explored the transmission of tokens during the computation process on devices through an Uncertainty Guided Sampling \cite{gan2022clouddevice} approach, aiming to enhance bandwidth utilization. However, this method exhibits strong randomness in the token selection, and the tokens selected may lack sufficient semantic information. Our UTS strategy allows us to reduce bandwidth while maximizing the semantic information richness of the selected image tokens.
In the cloud, knowledge distillation can be leveraged to transfer knowledge from large models to smaller models, aiming to minimize the parameter volume transmitted from the cloud to the device.  

Knowledge Distillation (KD) is a method of model compression and transfer learning~\cite{hinton2015distilling}. 
Over the years, many KD methods have been proposed that perform distillation over intermediate features \cite{huang2022masked, huang2023knowledge}, relation representation \cite{2020FKD, yang2022cross}, attention \cite{yang2022focal, zhang2023avatar} for various vision tasks. However, for MLLMs, there is currently no specific knowledge distillation method available to compress them effectively. In this paper, to better serve this system, we propose an adapter-based knowledge distillation(AKD) to get the manifolds embedded in multi-modal space.

\textbf{Continual Domain Adaptation.}
When devices are deployed in the real world, the continuous variation of data in real-world scenarios poses significant requirements for the generalization capability of models. 
To achieve better generalization performance on target data without access to source data, TENT \cite{wang2021tent} optimizes the pre-trained model's Batch Normalization layers through entropy minimization, while SHOT \cite{liang2021really} utilizes both entropy minimization and a diversity regularizer for information maximization. References \cite{li2020model} and \cite{liu2021sourcefree}  enhance model performance in target domains without source data by generating target-style data. 
Our work proposes a Cloud-Device Collaborative Continual Adaptation framework enables models to adapt to dynamically changing data distributions, significantly enhancing the generalization capability of device models. 
\section{Demo Video and Dataset}
We provide a video demo (the attached MP4 file), which contains the motivation and intuitive introduction of our proposed CD-CCA paradigm, the workflow of the overall framework, and the visualization results. Furthermore, the domain-shift VQAv2-to-AOKVQA-IDS dataset has been made available on Google Drive for researchers to access and utilize.

%


\end{document}